\documentclass[letterpaper, 10 pt, conference]{./style/IEEEconf}

\usepackage[pdftex]{graphicx}
\graphicspath{{images-src/}{images-bin/}} %

\usepackage{hyperref}
\usepackage{amsmath}
\usepackage{bm}
\usepackage{amssymb}
\usepackage{xspace}
\usepackage{mathtools}

\usepackage{booktabs}
\usepackage{subcaption}
\usepackage{braket}

\usepackage{cleveref}
\Crefname{equation}{Eq.}{Eqs.} %
\Crefname{figure}{Fig.}{Figs.}

\usepackage[textsize=scriptsize]{todonotes}
\usepackage[font=small]{caption}

\definecolor{BrickRed}{cmyk}{0, .89, .5, 0}
\definecolor{Blue}{cmyk}{.8, .3, .2, 0}
\definecolor{Green}{cmyk}{1, 0.2, 1, 0}
\definecolor{Black}{cmyk}{1, 1, 1, 0}

\definecolor{aliceblue}{rgb}{0.94, 0.97, 1.0}
\definecolor{almond}{rgb}{0.94, 0.87, 0.8}
\definecolor{amber}{rgb}{1.0, 0.49, 0.0}
\definecolor{amber}{rgb}{1.0, 0.49, 0.0}

\definecolor{bleu}{HTML}{04D9E0}
\definecolor{oran}{HTML}{F5AC44}
\definecolor{vert}{HTML}{BED95F}

\definecolor{GithubBlue}{HTML}{1c88e3}
\definecolor{GithubGreen}{HTML}{7baf44}
\definecolor{GithubRed}{HTML}{e22d2d}
\definecolor{LightGrey}{HTML}{F0F0F0}
\definecolor{CPGRed}{HTML}{a61e4d}
\definecolor{RLBlue}{HTML}{1864ab}

\definecolor{lilac}{HTML}{862e9c}
\definecolor{green}{HTML}{5c940d}
\definecolor{orange}{HTML}{d9480f}
\definecolor{violet}{HTML}{5f3dc4}
\definecolor{turquoise}{HTML}{0b7285}

\DeclareRobustCommand{\TODOCRITICAL}[1]{
\todo[backgroundcolor=red!20,inline]{\textcolor{BrickRed}{\the\value{section}.\the\value{subsection}) #1}}
}

\DeclareRobustCommand{\sidenote}[1]{
\todo{\the\value{section}.\the\value{subsection}) #1}
}

\newcommand{\eg}{e.\,g.\ }
\newcommand{\ie}{i.\,e.\ }
\newcommand{\cf}{cf.}

\newcommand{\bert}{\textit{bert}\xspace}

\newcommand{\E}[2]{\operatorname{\mathbb{E}}_{#1}\left[#2\right]}

\newcommand{\lab}[1]{\text{#1}}
\newcommand{\radius}{\rho}
\newcommand{\phase}{\varphi}
\newcommand{\gndclearance}{\Delta z_\lab{clear}}
\newcommand{\gndpenetration}{\Delta z_\lab{pen}}
\newcommand{\steplen}{\Delta x_\lab{len}}
\newcommand{\couplingweight}{c}

\newcommand{\parameters}{\alpha}
\newcommand{\bbofn}{f}

\DeclareMathOperator*{\argmin}{arg\,min}

\newcommand{\drift}{\Delta_{xy}}
\newcommand{\zvelreward}{|\dot{z}|}
\newcommand{\robotrollvel}{\dot{\psi}_\lab{roll}}
\newcommand{\robotyawvel}{\dot{\psi}_\lab{yaw}}

\newcommand{\stiffness}{k}
\newcommand{\torque}{\tau}
\newcommand{\motorpos}{\theta}
\newcommand{\motorvel}{\dot{\motorpos}}
\newcommand{\jointpos}{q}
\newcommand{\jointvel}{\dot{\jointpos}}

\newcommand{\stiffnessunit}{\text{Nm}/\text{rad}}

\newcommand{\ratio}{\mathcal{R}^\frac{\jointvel}{\motorvel}}

\newcommand{\state}{\mathbf{s}}
\newcommand{\reward}{r}

\newcommand{\st}{{\state_t}}
\newcommand{\rt}{{\reward_t}}

\newcommand{\policy}{\pi}

\newcommand{\tqc}{\textsc{TQC}\xspace}

\newcommand{\annotatedimage}[5]{
\begin{tikzpicture}
  \node[anchor=south west,inner sep=0] (image) at (0,0) {
    \includegraphics[#1]{#2}
  };
  \begin{scope}[x={(image.south east)},y={(image.north west)}]
    \node[black,text opacity=1,fill=white,fill opacity=0.5] at (#3, #4) {#5};
  \end{scope}
\end{tikzpicture}%
}

\newcommand{\fourgaitpic}[2]{%
\annotatedimage%
	{width=0.23\linewidth}{#1}%
	{0.90}{0.14}{\footnotesize #2}%
}

\newcommand{\multiannotatedimage}[2]{
	\begin{tikzpicture}[%
	every node/.style={draw=black, black, text opacity=1, fill=white, fill opacity=0.6,inner sep=0.5mm},%
	]
  \node[draw=white, anchor=south west,inner sep=0.5mm] (image) at (0,0) {
    #1
  };
  \begin{scope}[x={(image.south east)},y={(image.north west)}]
    #2
  \end{scope}
	\end{tikzpicture}%
}

\IEEEoverridecommandlockouts
\title{\LARGE \textbf{Learning to Exploit Elastic Actuators for Quadruped Locomotion}}

\author{Antonin Raffin$^1$, Daniel Seidel$^1$, Jens Kober$^2$, Alin Albu-Sch{\"a}ffer$^1$, Jo\~ao~Silv\'erio$^1$, Freek Stulp$^1$
	\thanks{$^1$ German Aerospace Center (DLR), Robotics and Mechatronics Center (RMC), M\"unchner Str. 20, 82234 We\ss ling, Germany}
	\thanks{$^2$ Delft University of Technology, Cognitive Robotics Department, The Netherlands}
}

\begin{document}

\maketitle

\begin{abstract}
Spring-based actuators in legged locomotion provide energy-efficiency and improved performance, but increase the difficulty of controller design.
While previous work has focused on extensive modeling and simulation to find optimal controllers for such systems, we propose to learn model-free controllers directly on the real robot.
In our approach, gaits are first synthesized by central pattern generators (CPGs), whose parameters are optimized to quickly obtain an open-loop controller that achieves efficient locomotion.
Then, to make this controller more robust and further improve the performance, we use reinforcement learning to close the loop, to learn corrective actions on top of the CPGs.
We evaluate the proposed approach on the DLR elastic quadruped \bert.
Our results in learning trotting and pronking gaits show that exploitation of the spring actuator dynamics emerges naturally from optimizing for dynamic motions, yielding high-performing locomotion, particularly the fastest walking gait recorded on \bert, despite being model-free.
The whole process takes no more than 1.5 hours on the real robot and results in natural-looking gaits.
\end{abstract}

\section{Introduction}
\begin{figure}[ht]
	\centering
	\includegraphics[width=\linewidth,trim={0 25 0 0},clip]{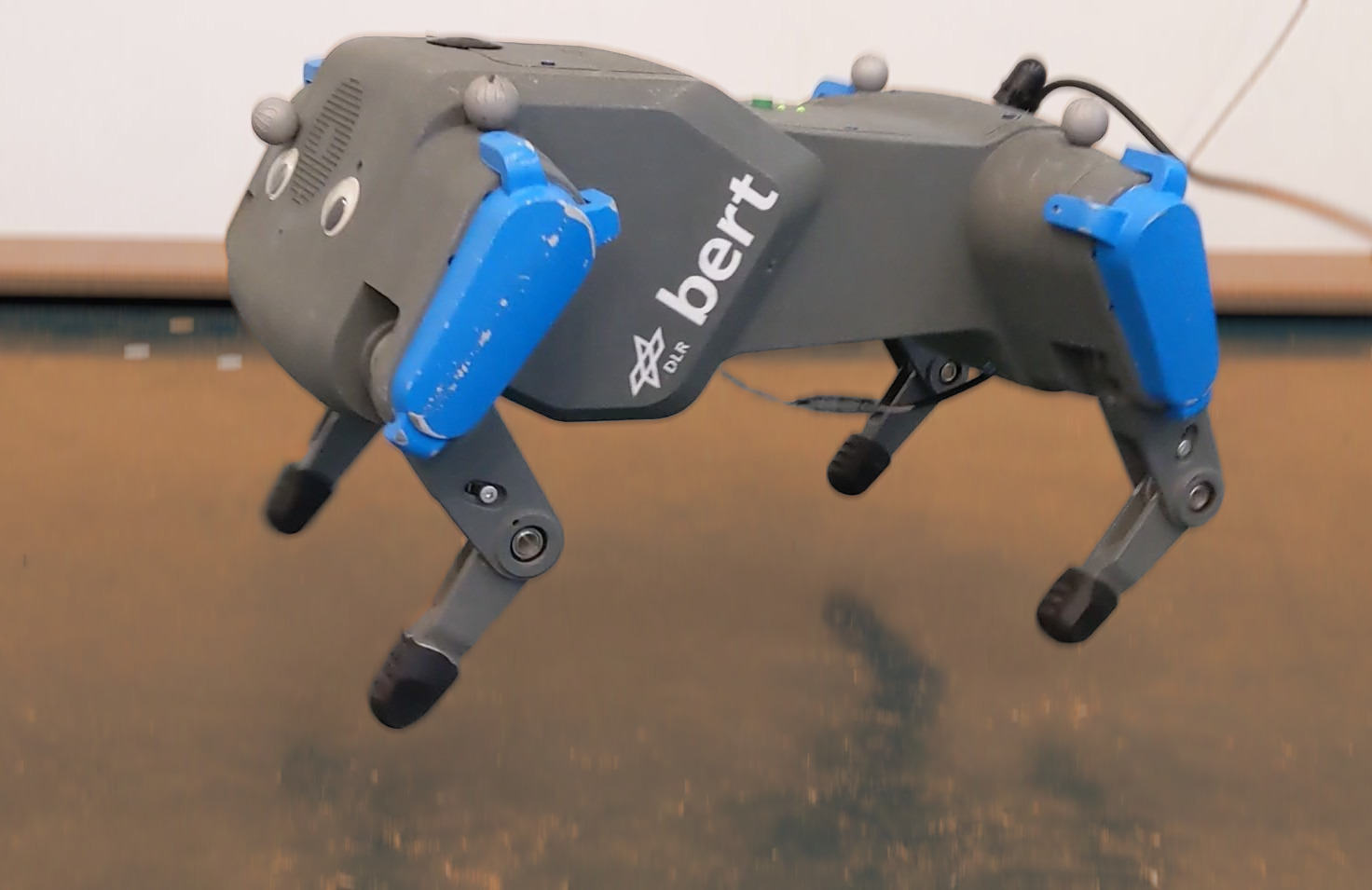}
	\caption{Compliantly actuated DLR quadruped \bert jumping in place.}
	\label{fig:bert}
\end{figure}

Robots with elastic actuators, such as DLR's \bert~\cite{seidel2020rough} (\Cref{fig:bert}), are a promising alternative to their rigid counterparts~\cite{mansfeld2021speed, bjelonic2023learning}, especially for legged locomotion~\cite{alexander1990springs, grimmer2012comparison,ruppert2022learning,kormushev2019learning}.
In addition to the passive compliance that protects them from collisions, the elastic elements provide energy storing and improved efficiency if properly used.
The design of controllers that fully exploit the intrinsic dynamics of compliantly actuated robots remains a challenge, requiring high amounts of time and system-specific knowledge~\cite{keppler2018elastic}.
In this work, we address the problem of obtaining efficient controllers for the locomotion of elastically-actuated robots by formulating controller design as a learning problem.
\subsection{Related Work}
\textbf{Quadruped robots.} In recent years, several works studied quadruped locomotion controllers on different hardware platforms (ANYmal, A1 Unitree, \ldots).
Even though some of these have elastic actuators (\eg ANYmal~\cite{hutter2012starleth} or Mini Cheetah~\cite{katz2019mini}), their stiffness is high compared to \bert (see \Cref{sec:bert}) leaving little room to exploit the advantages of the actuators' elasticity in control.

\textbf{Reinforcement learning for locomotion.} Obtaining locomotion controllers is commonly based on fast simulators and massive parallelism~\cite{miki2022learning, rudin2022learning} and consists of learning the controllers in simulation and subsequently deploying them on real hardware~\cite{lee2020learning, kumar2021rma}.
Learning in simulation, however, requires an accurate model of the robot~\cite{hwangbo2019learning}, complex reward engineering, and there is no guarantee that a controller working in simulation can be deployed on the real robot.
Heavy randomization and generation of diverse training environments help mitigate these issues~\cite{fu2021minimizing, rudin2022learning}, but still requires engineering efforts to find the right balance between environments too hard to solve and ones that do not transfer to the real robot.
Our key difference from many previous approaches is that we learn directly on the real hardware, without the need to transfer from simulation to reality.
To our knowledge, this is also the first time that reinforcement learning is used directly on a quadruped with elastic actuators.

\textbf{Training on real robot.} Thanks to more robust hardware~\cite{buchler2022learning} and more sample-efficient algorithms~\cite{chatzilygeroudis2019survey, smith2022walk, wu2022daydreamer, hiraoka2022dropout}, learning directly on the real hardware in only a few hours is now possible~\cite{calandra2016bayesian, raffin2021gsde}.
Despite these advances, training on real robots can nevertheless be time-consuming and often does not yield natural-looking gaits~\cite{smith2022walk, wu2022daydreamer}.
In the present work, we aim both at reducing the training time (and thus the robot wear and tear) and obtaining gaits that look natural by 1) providing domain knowledge in the form of central pattern generators and 2) exploiting the natural dynamics of the robot, in particular its elastic joints.
In order to mitigate model mismatch or simulation-to-reality gaps, we choose to apply a model-free learning approach directly on the real hardware.

\textbf{CPGs and Reflex-Based Controllers.} Inspired by biology, central pattern generators (CPGs) have been proposed for locomotion~\cite{crespi2008online, righetti2008pattern, ajallooeian2013central,bellegarda2022cpgrl}.
They generate gaits that look natural but require careful parameter tuning, usually done in simulation.
In this work, we rely on a CPG model to represent the open-loop motion policy that is later improved with reinforcement learning.
Another possibility to obtain dynamic gaits in elastically-actuated robots is to rely on reflex-based controllers~\cite{seidel2020rough, lakatos2018dynamic, schmidt2021adapting}, whose parameters also require further optimization.
Our research is applicable to both CPG and reflex-based controllers, as we automate the parameter optimization and improve robustness and performance with a learned reactive controller.
Our approach achieves the fastest walking gait on the \bert hardware (0.34~m/s vs 0.25~m/s in~\cite{lakatos2018dynamic}).

\textbf{Comparison to CPG-RL and PMTG.} Parallel to our work is the CPG-RL method~\cite{bellegarda2022cpgrl}, which can be viewed as a special case of Policies Modulating Trajectory Generators (PMTG)~\cite{iscen2018policies} where the residual component is removed.
In CPG-RL and PMTG, the parameters of the trajectory generator are dynamically updated by an RL agent.
Both approaches focus primarily on simulation and therefore rely on domain randomization to ensure performance during real-world testing.
Despite sharing architectural similarities, unlike CPG-RL and PMTG, our research emphasizes the unique challenges and benefits associated with learning controllers for quadruped robots with elastic actuators.
While CPG-RL employs a massively parallel simulator with 4096 virtual robots, we train directly on a single real robot, concluding that its elastic actuators indeed provide additional robustness and performance over rigid equivalents.
Furthermore, we challenge the statement made in the PMTG paper that ``\textit{despite the pre-optimization, the TG itself cannot provide stable forward locomotion since it lacks the feedback from the robot}'':
our results show that an optimized open-loop controller is sufficient for forward motion on the real hardware.

\textbf{Feedforward and Feedback Control.} The integration of feedforward (open-loop) and feedback (closed-loop) control is a well-established practice in engineering~\cite{goodwin2000control,astrom2008feedback,della2017controlling}, also known as \textit{residual learning}~\cite{johannink2019residual} in the RL community.
This approach combines the benefits of fast computation and robustness to sensor noise offered by open-loop control with the ability of closed-loop control to address model inaccuracies and respond to disturbances.
In our work, we adopt this architecture by utilizing a bio-inspired open-loop control strategy based on central pattern generators (CPGs) and augmenting it with reinforcement learning to enhance performance and robustness.
Importantly, \textit{both components of our control system are model-free and optimized directly on the real robot, enabling us to fully leverage the intrinsic dynamics of the elastic robot}.

\subsection{Contributions}
In summary, the main contributions of our paper are:
\begin{itemize}
	\item an approach to design locomotion controllers for robots with elastic actuators (\Cref{sec:learning}) that combines CPGs, black-box optimization and reinforcement learning, achieving the fastest walking gait for the \bert platform,
	\item an experimental verification that the learned controllers exploit the elasticity of the actuators (\Cref{sec:results}), and
	\item showing that controllers can be learned directly on the real elastic robot in less than one hour, without the need for complex reward engineering or massively parallel simulation~\cite{rudin2022learning} (\Cref{sec:results}).
\end{itemize}

\section{Background}

In this section, we provide an overview of DLR's elastic quadruped \bert, metrics for monitoring spring usage, as well as central pattern generators -- a cornerstone of our proposed approach -- including the CPG parameters that we aim to optimize.

\subsection{DLR quadruped \bert}
\label{sec:bert}
\begin{figure}[ht]
	\centering
	\includegraphics[width=0.3\linewidth]{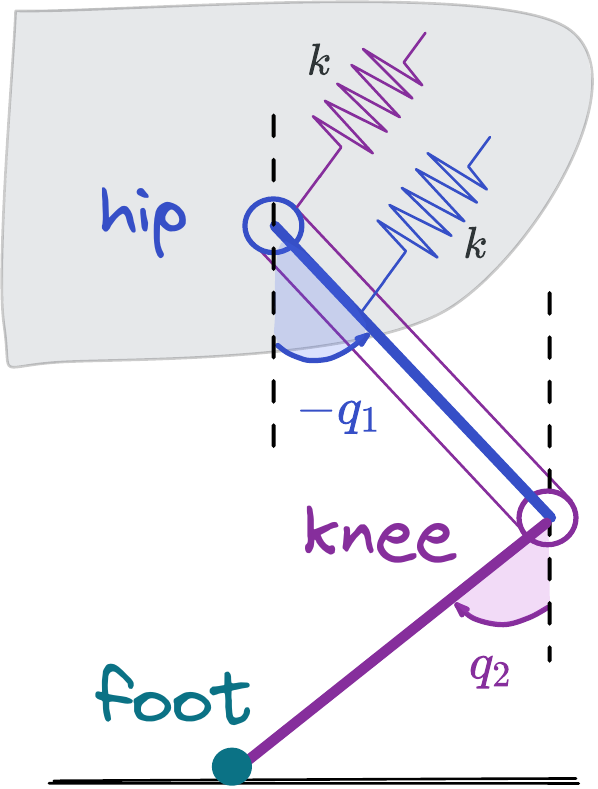}\hfill
	\includegraphics[width=0.5\linewidth]{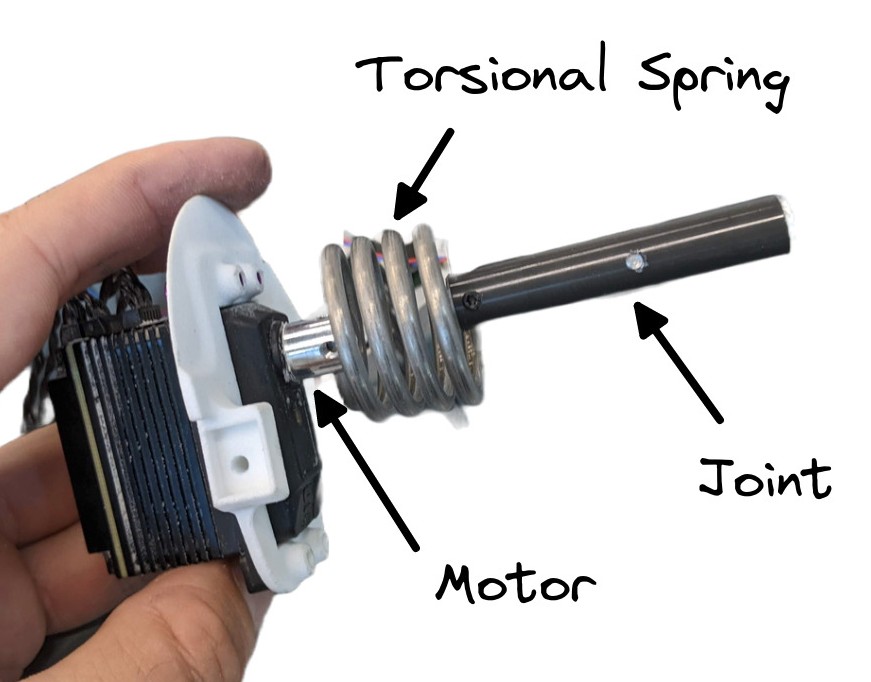}
	\caption{Model of the compliantly actuated leg (left) and a serial elastic actuator (right).}
	\label{fig:leg}
\end{figure}

The DLR quadruped \bert~\cite{seidel2020rough} (shown in~\Cref{fig:bert}) is a cat-sized quadruped robot with series elastic actuators (SEA).
The legs are attached to the trunk through the hip axis, while the motors are located inside the trunk and connected by belt drives.
Each leg consists of a hip and a knee joint and is composed of two segments of equal length of approximately 8cm. %
The total mass of the robot is about 3.1kg. %
As depicted in \Cref{fig:leg}, motors are connected to the links via a linear torsional spring with constant stiffness $\stiffness \approx 2.75 \stiffnessunit$. We use the deflection of the spring to estimate the external torque applied to each joint:
\begin{equation}
	\label{eq:torque}
	\torque_i = \stiffness (\motorpos_i - \jointpos_i)
\end{equation}
where $\motorpos$ is the motor position before the spring and $\jointpos$ the joint position.

\bert is equipped with very soft springs relative to its weight and lever arm, \ie $2.75 \stiffnessunit$, compared to for instance
StarlETH~\cite{hutter2012starleth} with 20cm link length, total weight of 23kg and $\stiffness {=} 70 \stiffnessunit$.
This means that \eg when standing in a default position with joints at $30 \deg$, \bert can be pushed into joint limits without motor movement, while StarlETH can be pushed into its springs only minimally.
This makes \bert suitable for exploiting its springs as energy storage: forces required to load the springs are quite low and occur during normal gait motions.

When released at the right time, the stored energy can be converted into kinetic energy (joint velocity).
This can lead to improved task performance.
We formally define metrics to quantify the spring usage in the next section.

\subsection{Metrics}
Objective functions of learning methods typically describe tasks (\eg move forward as fast as possible),
and do not enforce explicit usage of the compliant actuators.
To measure if the learned controllers exploit the capability of the elastic robot, we define  metrics that quantify the spring usage.

By storing and releasing energy, springs allow joints to reach much higher velocities than the motors~\cite{mansfeld2021speed}.
The peak joint velocity usually has a delay compared to the peak motor velocity.
Therefore, we propose to compare the ratio between maximum joint and motor velocity over a trajectory for each joint:
\begin{equation}
	\label{eq:ratio}
	\ratio_i = \frac{\max_t{\jointvel_i}}{\max_t{\motorvel_i}}.
\end{equation}

Intuitively, if the springs are fully exploited, the ratio $\ratio$ should be greater than one (rigid robot baseline), meaning that the energy is stored and released at the right moment.

Finally, we also monitor the potential and kinetic energy to better understand how energy is transferred and how the learned controllers utilize the elastic actuators.
We only consider the kinetic energy related to the task (movement of the center of mass) and therefore do not plot the rotational kinetic energy.

\subsection{Central pattern generators (CPGs) for legged locomotion}

CPGs are a common tool used for generating open-loop policies for locomotion~\cite{ijspeert2008cpg}.
We rely on an implementation based on a system of four coupled nonlinear oscillators~\cite{righetti2006dynamic,righetti2008pattern,ajallooeian2013central} (one per leg) that have phase-dependent frequencies.
The equation of one oscillator in polar coordinates is~\cite{ijspeert2007swimming}:
\begin{equation}
	\begin{aligned}
		\label{eq:hopf}
		\dot{\radius_i} &= a (\mu - \radius_i^2)\radius_i \\
		\dot{\phase_i} &= \omega + \sum_j \, \radius_j \, \couplingweight_{ij} \, \sin(\phase_j - \phase_i - \Phi_{ij}) \\
		\omega &= \begin{cases}
		  \textcolor{turquoise}{\omega_\lab{swing}} &\text{if $\sin(\phase_i) > 0$}\\
		  \textcolor{lilac}{\omega_\lab{stance}} &\text{otherwise}
		\end{cases}
	\end{aligned}
\end{equation}
where $\radius_i$ and $\phase_i$ are the amplitude and phase of oscillator $i$, $\omega$ is the frequency of oscillations in rad/s, $\sqrt{\mu}$ is the desired amplitude, $\textcolor{turquoise}{\omega_\lab{swing}}$ and $\textcolor{lilac}{\omega_\lab{stance}}$ are the frequencies of the swing and stance phases. $a$ is a positive constant that controls the rate of convergence to the limit cycle.
The weights $\couplingweight_{ij}$ and phase biases $\Phi_{ij}$ define the couplings between oscillators.

\begin{figure}[ht]
	\centering
	\includegraphics[width=\linewidth]{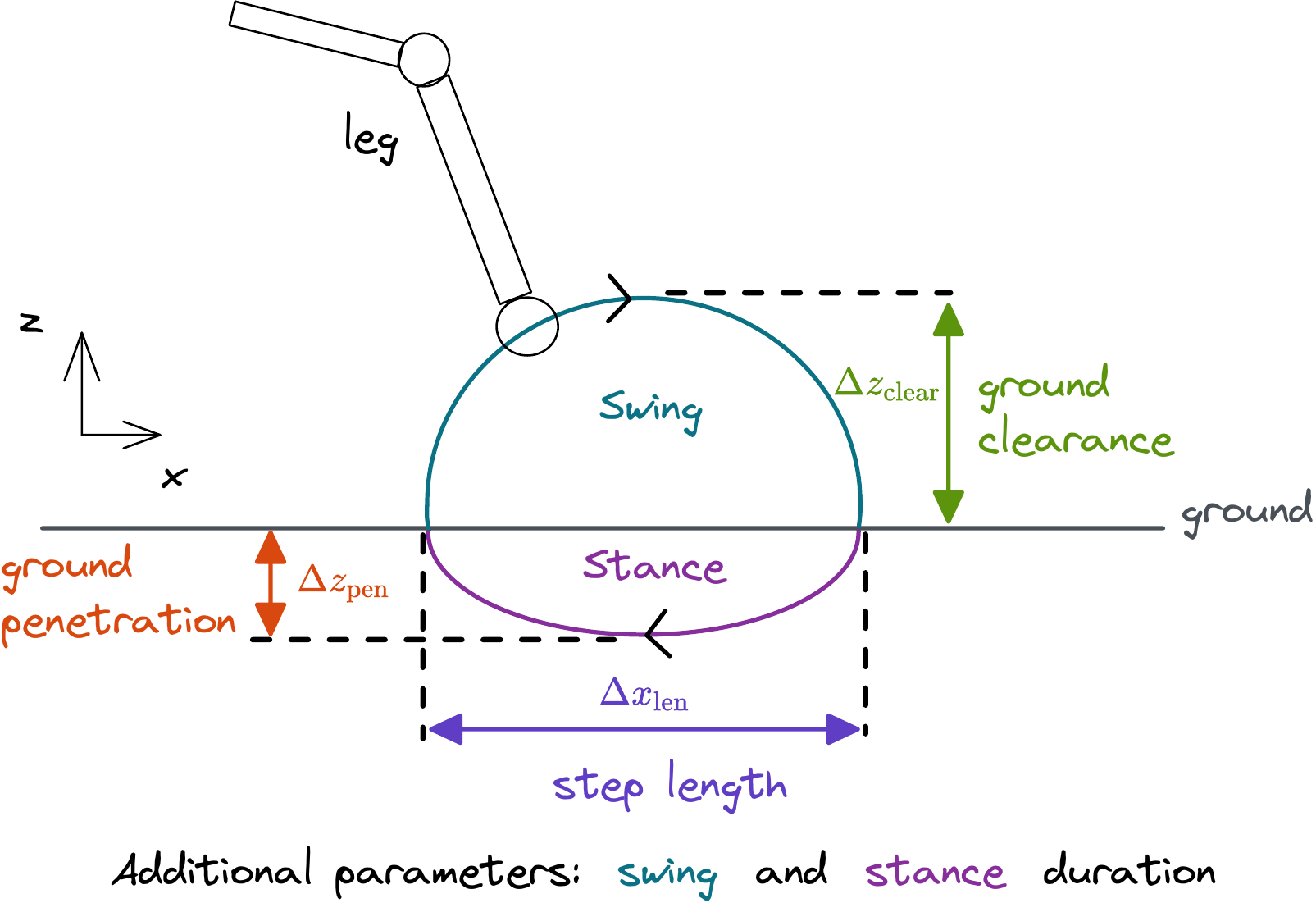}
	\caption{The parameters of the CPG gait that are optimized.}
	\label{fig:cpg}
\end{figure}

The output of each oscillator determines the foot trajectory in Cartesian space (\cf~\Cref{fig:cpg}).
The coupling matrix $\Phi$ therefore determines the gait.
For instance, for trotting~\cite{hildebrand1965gaits, bhatti2017gait} (foot pattern displayed in~\Cref{fig:pattern_trot}), the front right leg is in phase with the hind left leg, and they have a half-cycle offset with the other diagonal (front left with hind right).

As shown in~\Cref{fig:cpg}, the trajectory of each leg $i$ depends on four parameters that have to be tuned: ground clearance $\textcolor{green}{\gndclearance}$ and penetration $\textcolor{orange}{\gndpenetration}$, step length $\textcolor{violet}{\steplen}$, swing $\textcolor{turquoise}{\omega_\lab{swing}}$ and stance $\textcolor{lilac}{\omega_\lab{stance}}$ frequencies.
We convert the output of one oscillator $i$ to a desired foot trajectory following:
\begin{equation}
	\label{eq:cpg_traj}
	\begin{aligned}
		x_{\text{des},i} &= \textcolor{violet}{\steplen} \cdot \radius_i \cos(\phase_i)\\
		z_{\text{des},i} &= \Delta z \cdot \radius_i \sin(\phase_i)\\
		\Delta z &= \begin{cases}
		  \textcolor{green}{\gndclearance} &\text{if $\sin(\phase_i) > 0$ (\textcolor{turquoise}{swing})}\\
		  \textcolor{orange}{\gndpenetration} &\text{otherwise (\textcolor{lilac}{stance}).}
		\end{cases}
	\end{aligned}
\end{equation}
The most critical parameters for an elastically-actuated quadruped are the swing and stance duration, as they should match the natural dynamics of the robot~\cite{della2020using, albu2020review}.

Tuning these variables manually is time-consuming and usually results in suboptimal choices.
Therefore, we automate their tuning using black-box optimization (BBO) and run the optimization directly on the real robot, avoiding any model mismatch.
In the following section, we explain the use of BBO and how it fits into our proposed framework.

\section{Online learning for legged locomotion}
\label{sec:learning}

\begin{figure}[ht]
	\centering
	\includegraphics[width=\linewidth]{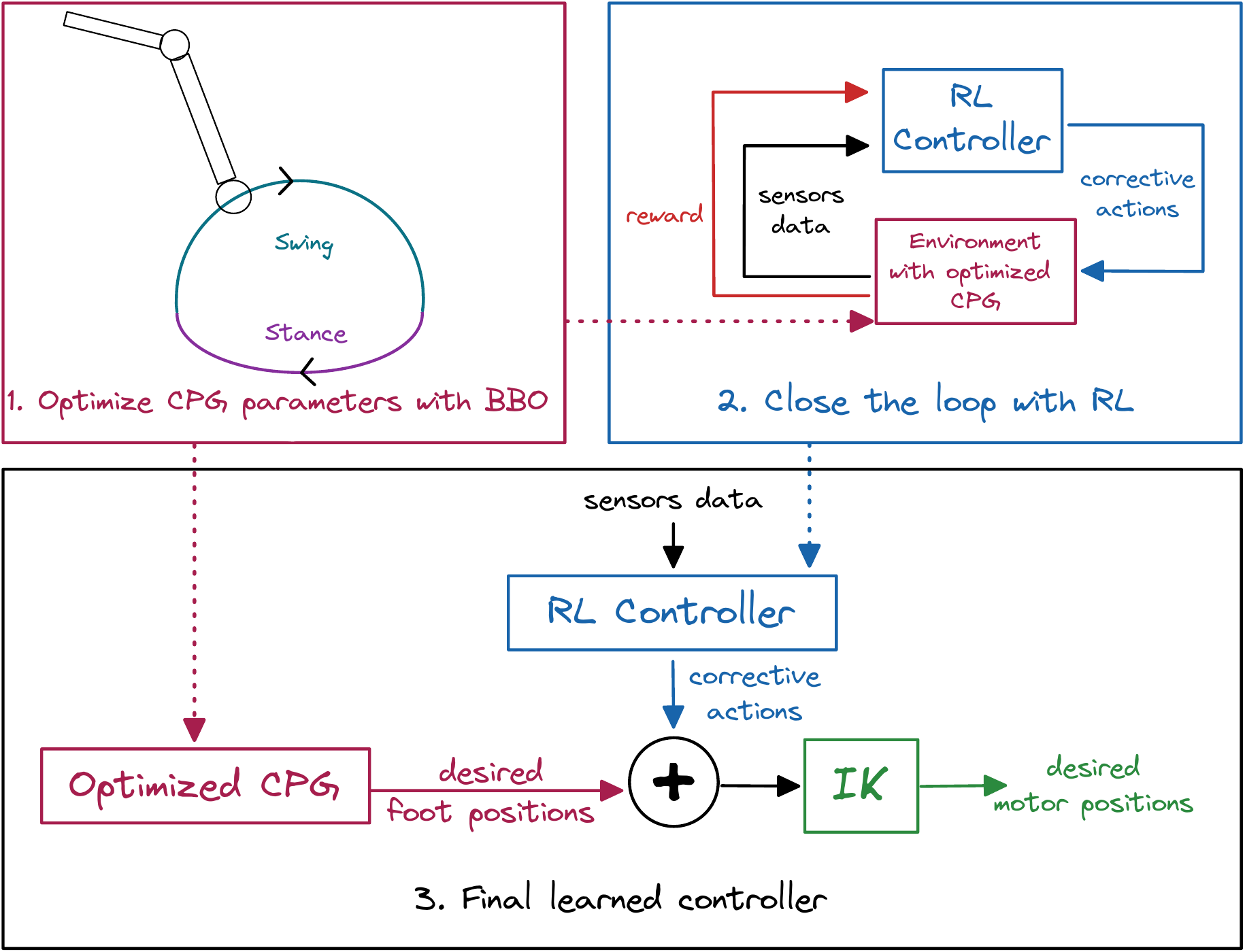}
	\caption{Overview of the proposed approach: CPG parameters are first optimized using BBO, then a RL controller is trained on top and outputs corrective actions.}
	\label{fig:overview}
\end{figure}

To learn a locomotion controller for the elastic legged robot, we combine two model-free methods.
We first optimize the parameters of a selected gait (generated from a CPG) to quickly obtain an open-loop controller that works in the nominal case without disturbances.
Then, we refine the controller and make it more robust to perturbations by learning a reinforcement learning (RL) controller that adds offsets to the CPG output.
The entire optimization process -- depicted in~\Cref{fig:overview} -- is conducted directly on the real robot, removing the need for modeling, accurate simulators or simulation-to-reality transfer.

CPGs provide a compact representation of limb end-effector trajectories (\Cref{eq:cpg_traj}) that can be adjusted via the set of parameters $\parameters$ shown in~\Cref{fig:cpg}:
\begin{equation}
	\parameters = \set{\textcolor{green}{\gndclearance}, \textcolor{orange}{\gndpenetration}, \textcolor{violet}{\steplen}, \textcolor{turquoise}{\omega_\lab{swing}}, \textcolor{lilac}{\omega_\lab{stance}}}.
\end{equation}

\subsection{Open-loop policy learning with black-box optimization (BBO)}
In BBO, the goal is to find a set of parameters $\parameters \in \mathbb{R}^n$ that minimizes an objective function $\bbofn:\mathbb{R}^n\rightarrow\mathbb{R}$,

\begin{equation}
	\label{eq:bbo_obj}
	\parameters^* = \argmin_{\parameters}{\bbofn(\parameters)}.
\end{equation}
A BBO algorithm only has access to function evaluations of $\bbofn$.
It treats $\bbofn$ as a black box and does not make any assumption about it (\eg $\bbofn$ can be non-differentiable).
In our case, each function evaluation corresponds to one trial (or episode) on the robot for a given set of CPG parameters $\parameters$.

We use $\bbofn$ to quantify how much a set of CPG parameters satisfies a desired goal.
We will describe the objective functions we choose for each task in~\Cref{sec:task_spec}.

\subsection{Closing the loop with reinforcement learning}
\label{sec:closeloop}
Reinforcement learning (RL) is similar to BBO but has access to the immediate reward $\reward_t$ that composes the objective function, when it's available, making it more sample-efficient.
Formally, an agent interacts with its environment and generates trajectories $\zeta_\policy$ that follow a policy $\policy$.
By progressively updating its policy, the agent learns to maximize its cumulative (discounted) reward:
\begin{equation}
	\label{eq:rl_obj}
  \sum_t \E{\zeta_\policy}{\gamma^t \reward_t}
\end{equation}
where $\gamma \in [0,1)$ is the discount factor and represents a trade-off between maximizing short-term and long-term rewards.

While the BBO algorithm optimizes the parameters of the gait generated by the coupled oscillators, the \textcolor{RLBlue}{RL} agent learns a reactive controller $\pi$ as a set of \textcolor{RLBlue}{corrective actions} per leg $i$ defined as $\textcolor{RLBlue}{\bm{\pi}_i=\left[\pi_{x,i}\left(\st\right), \pi_{z,i}\left(\st\right)\right]}$.
The learned policy $\pi$ adjusts \textcolor{CPGRed}{desired foot positions} produced by the \textcolor{CPGRed}{CPG} (\Cref{eq:cpg_traj}) given the current state of the robot $\st$, closing the control loop:

\begin{equation}
	\label{eq:cpg_rl}
	\begin{aligned}
		x_{\text{des},i} &= \textcolor{CPGRed}{\steplen \cdot \radius_i \cos(\phase_i)} + \textcolor{RLBlue}{\pi_{x,i}(\st)} \\
		z_{\text{des},i} &= \textcolor{CPGRed}{\Delta z \cdot \radius_i \sin(\phase_i)} + \textcolor{RLBlue}{\pi_{z,i}(\st)}
	\end{aligned}
\end{equation}

\subsection{Task Specification}
\label{sec:task_spec}

In this work, we focus on two dynamic gaits: trotting and pronking (jumping in place).
\paragraph{Trotting}
The first task we are interested in is to obtain a fast walking trot,
where we optimize for mean forward speed ($\bbofn(\parameters) = - \frac{\Delta x}{\Delta t}$).
We define the immediate reward as the distance traveled between two timesteps along the desired axis ($\rt = \dot{x}_\lab{robot}$).

After optimization, we further evaluate the five best performing candidates for more episodes.
This post-processing step is key to filtering out the evaluation noise, \ie it helps to find parameters that lead to reliable gaits.

\begin{figure}[ht]
	\centering
	\includegraphics[width=0.95\linewidth,trim={0 15 0 0},clip]{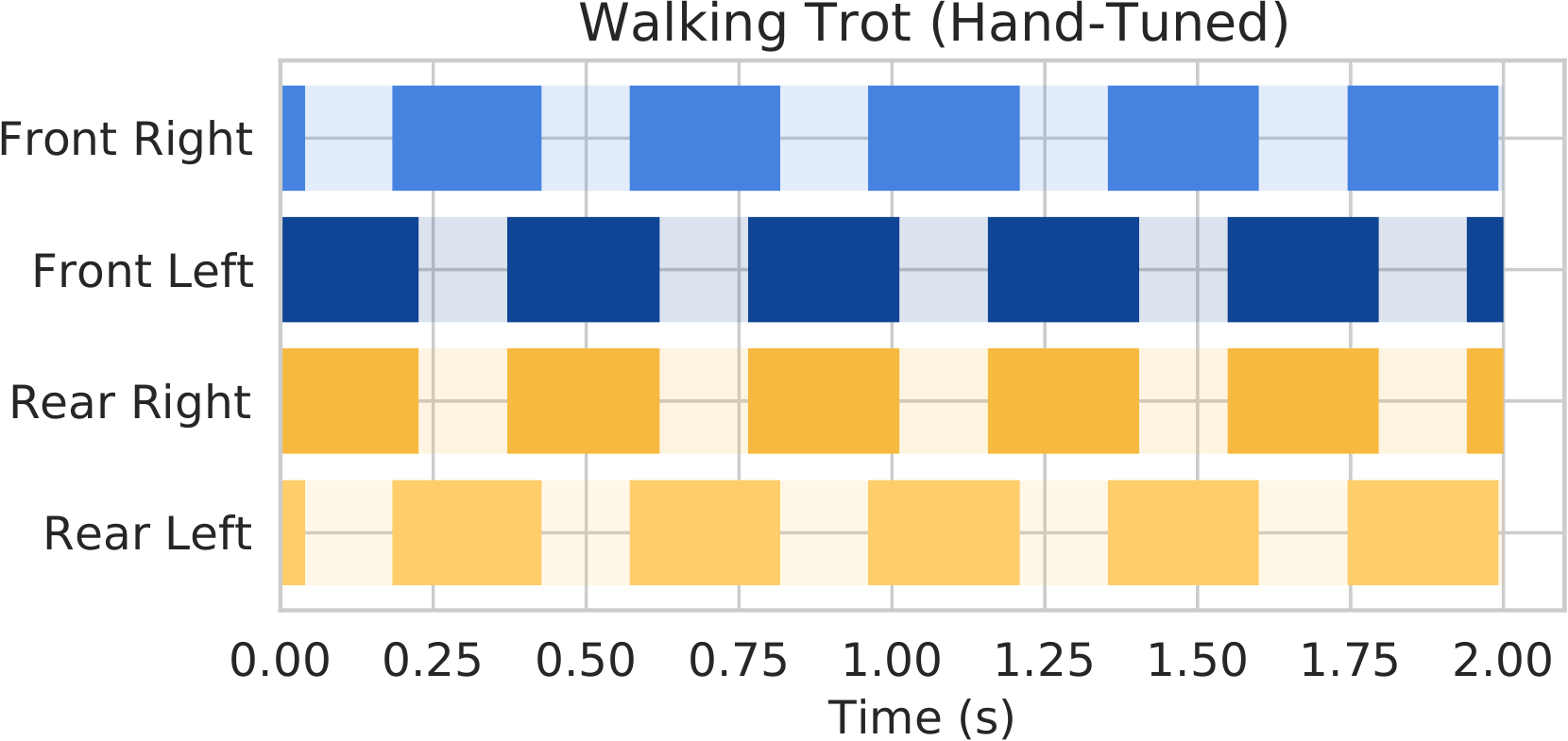}
	\includegraphics[width=0.95\linewidth]{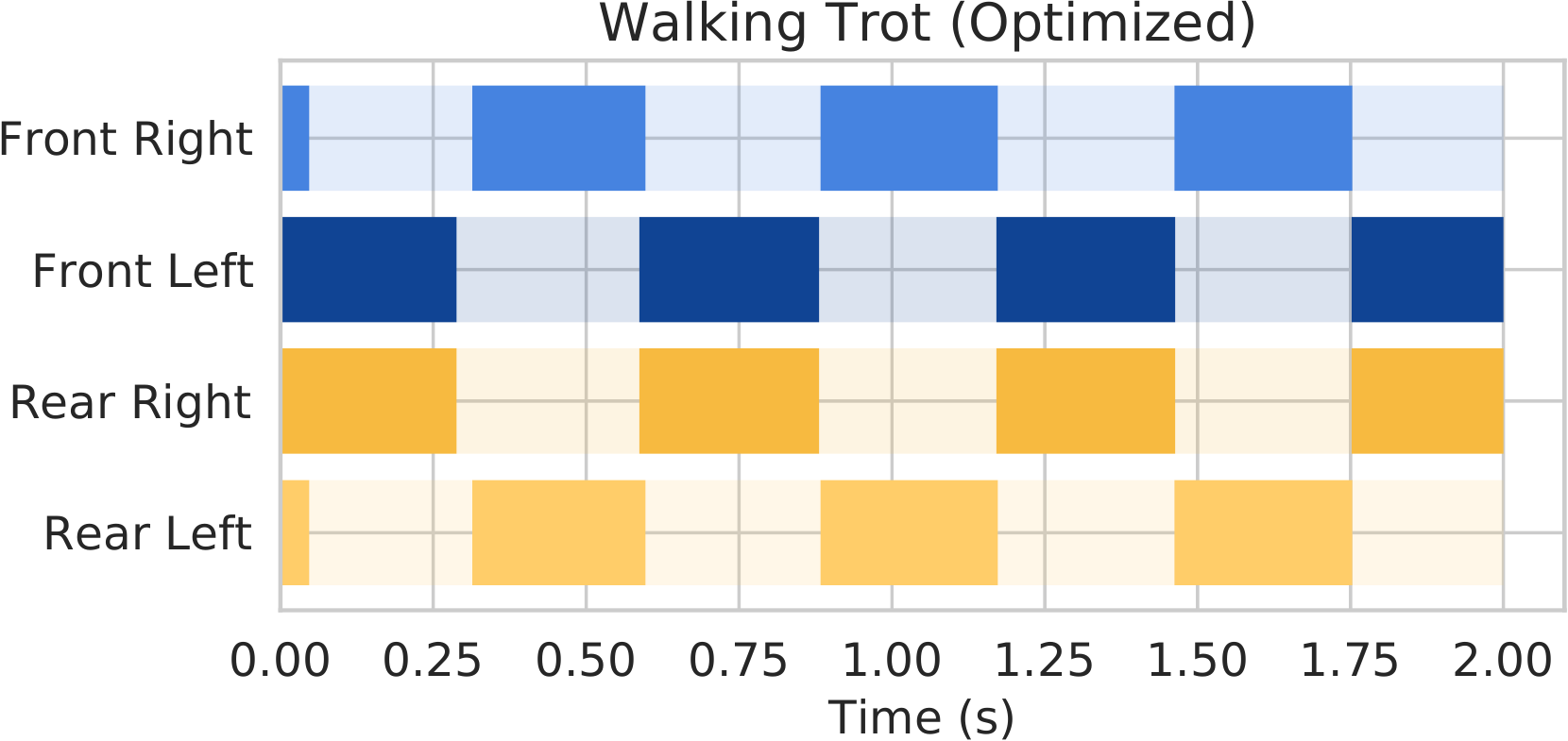}
	\caption{Trotting gaits for a 2-second period.
	Dark colors represent stance phases, light colors swing phases.}
	\label{fig:pattern_trot}
\end{figure}

\begin{figure}[ht]
  \centering
  \includegraphics[width=0.95\linewidth,trim={0 15 0 0},clip]{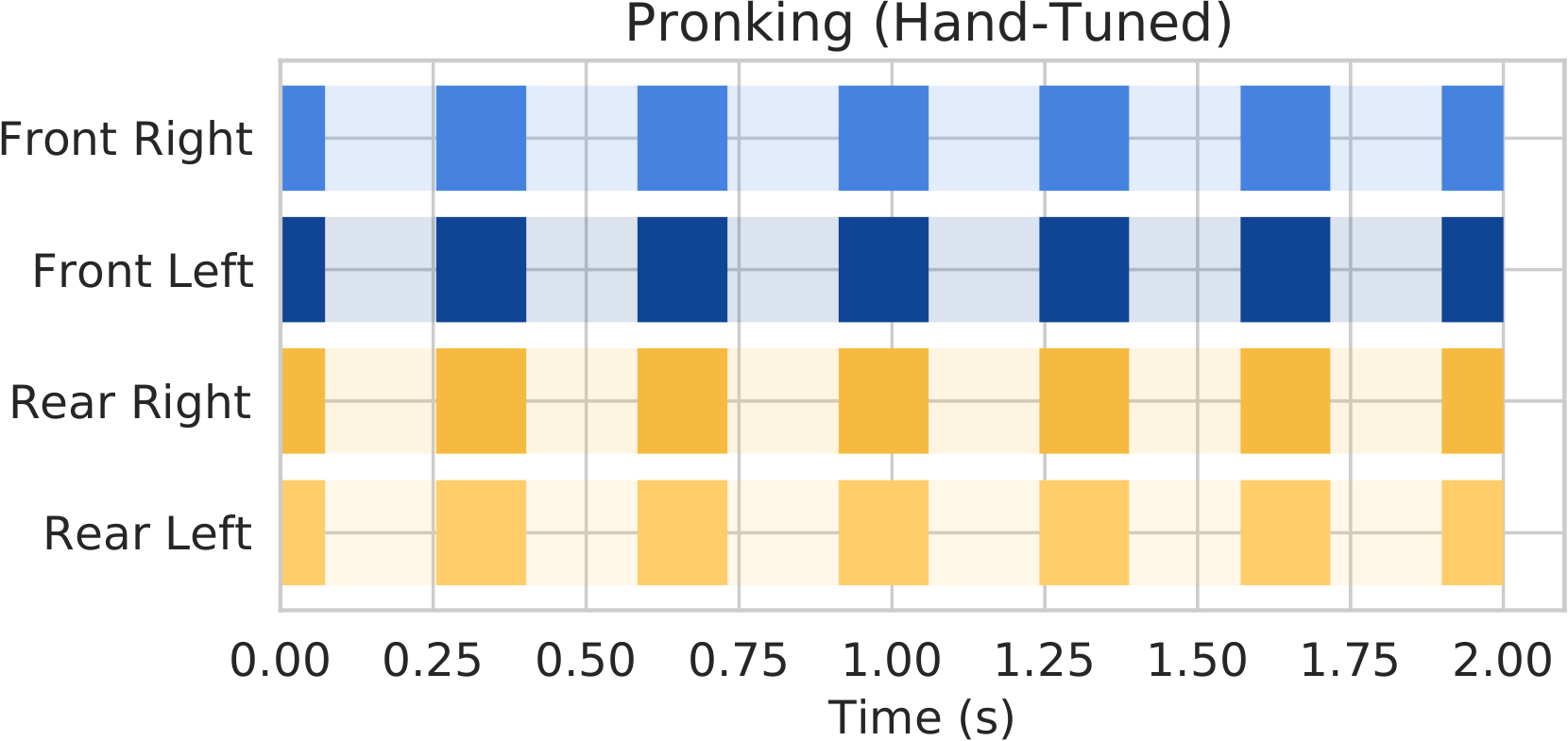}
  \includegraphics[width=0.95\linewidth]{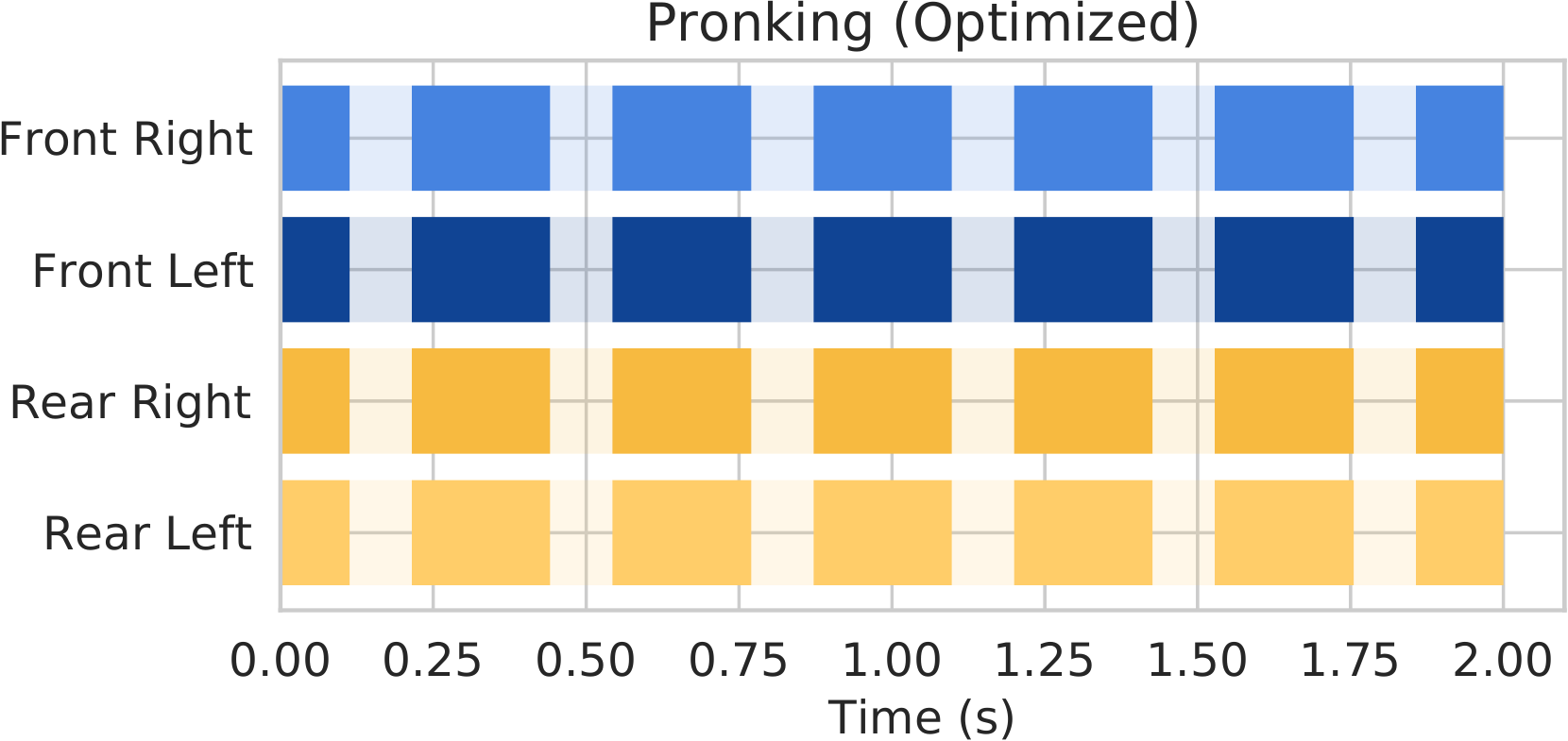}
\caption{Pronking gaits for a 2-second period.}
\label{fig:pattern_pronk}
\end{figure}

\paragraph{Pronking}
The second task is to jump in place (pronk) without falling or drifting from the starting position.
In a pronking gait, all legs are synchronized: they all take off and land at the same time (as seen in~\Cref{fig:pattern_pronk}).
The step length is set to zero in that case.
We penalize angular velocity for roll and yaw, distance to the starting point and reward velocity along the $z$ axis (aligned with gravity field):
$\rt = w_1 \zvelreward - w_2 (\robotrollvel^2 + \robotyawvel^2) - w_3 \drift^2$, where $w_{1,2,3}$ are weights chosen in a way that the primary reward $\zvelreward$ has a higher magnitude than the secondary costs.
We define the objective function for the BBO as the total reward per episode ($\bbofn(\parameters) = -\sum_t \rt$).

In all tasks, the RL agent receives as input $\st$ the current joint positions, velocities, torques, linear acceleration and angular velocity (from the IMU), and the desired foot position generated by the CPG controller.
The action space corresponds to offsets for each foot in Cartesian space.

\section{Results}
\label{sec:results}

We evaluate our approach on two locomotion tasks (trotting and pronking) to answer the following questions:
\begin{itemize}
	\item Can we quickly learn a controller directly on the real elastic robot?
	\item How does the learned controller exploit the elastic actuators?
	\item What do we gain by learning CPG parameters with BBO?
	\item What is the added value of closing the loop with RL?
	\item Is CPG needed or is RL from scratch enough?
\end{itemize}

\subsection{Setup}
During training only, we use an external tracking system that provides robot absolute position and orientation at $60 \text{Hz}$.
Each trial has a timeout of $5 \text{s}$ and terminates earlier if the robot reaches the tracking limits or if the quadruped falls over.
We use a treadmill of one by three meters to limit the need of manual intervention.
The treadmill is also the tracked area usable by the robot.

To tune the parameters of the open-loop controller, we use
Bayesian optimization, as it is sample efficient in low-dimensional search spaces~\cite{calandra2016bayesian}. We use the TPE~\cite{bergstra2011tpe} implementation from the Optuna~\cite{takuya2019optuna} library.
For reinforcement learning we choose \tqc, a sample-efficient off-policy algorithm~\cite{kuznetsov2020tqc}, with smooth exploration~\cite{raffin2021gsde} to remove the need for a low-pass filter.
We run RL experiments using the RL Zoo~\cite{raffin2020rlzoo} framework and Stable-Baselines3~\cite{raffin2021sb3} library (with tuned hyperparameters from~\cite{raffin2021gsde}).

The agent is trained at $30 \text{Hz}$ but can be evaluated at $60 \text{Hz}$ (limited by the tracking system).
The CPG parameters are optimized during 45 minutes for the trotting experiment and 25 minutes for the pronking one (250 and 160 trials respectively).
The RL controller is then trained on top during one hour.
For safety, the motor velocities are capped at $\motorvel_{\max} = 4 \text{rad/s}$.

\textbf{Hand-tuned Baseline.} We use parameters tuned by hand by a human experimenter familiar with the elastic quadruped as a baseline.
The experimenter was given the same time and search space as the BBO algorithm, as well as an explanation of the meaning and importance of each parameter.

\textbf{RL Baseline.} When learning from scratch, we use the setting described in~\cite{smith2022walk} to be able to learn a controller in minutes, using 10 gradient steps per control step.
The action space is extended to be similar in task space to the one used by the CPG controller.

\subsection{Fast Trotting}

\begin{figure}[ht]
	\centering
	\includegraphics[width=\linewidth]{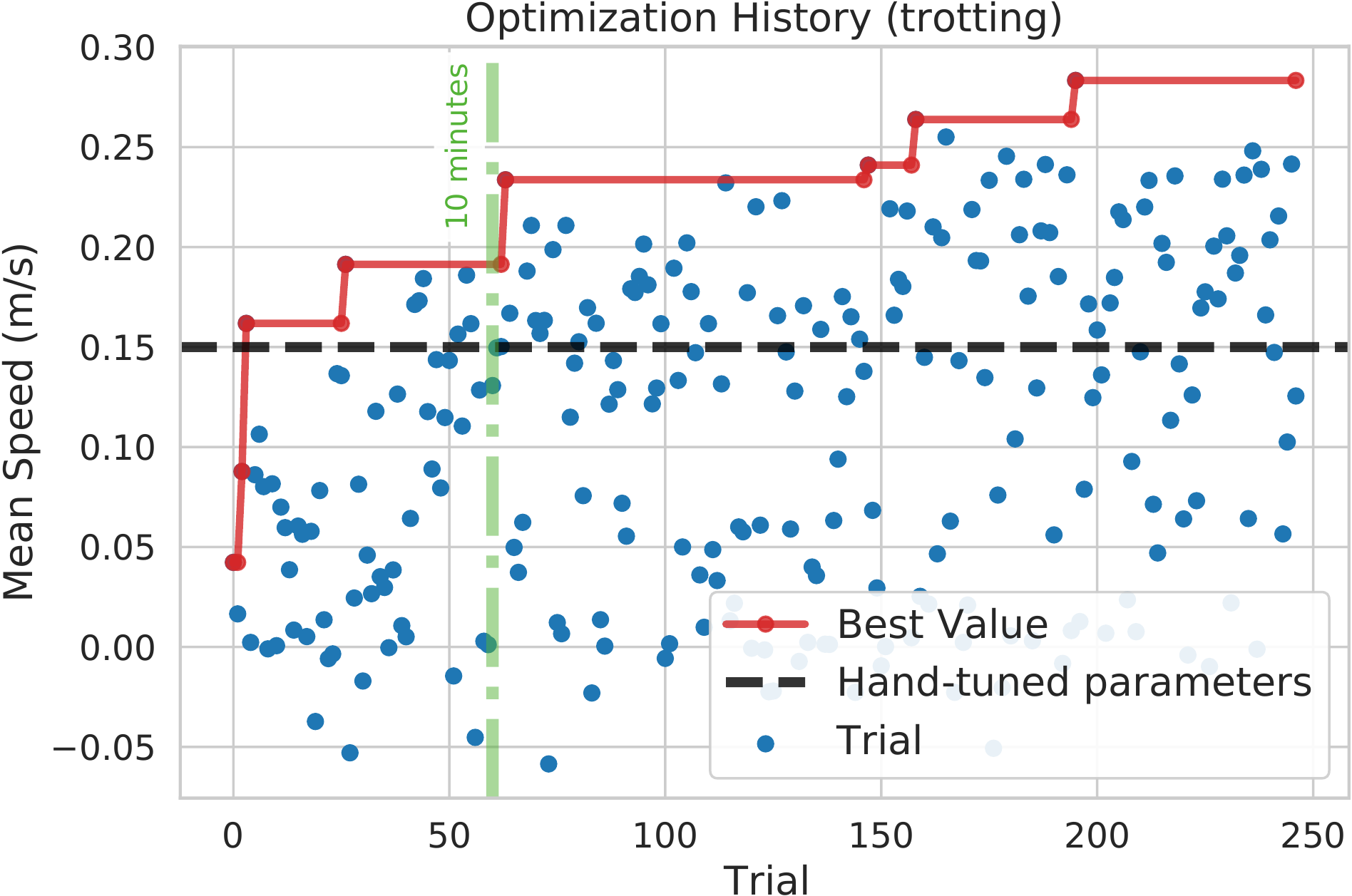}
	\caption{Optimization history plot for trotting (45 minutes).}
	\label{fig:optim_hist_trot}
\end{figure}

Without the need for a simulator or prior knowledge, the automatic tuning with BBO quickly finds good parameters for trotting in less than an hour with the real robot.
Looking at the optimization history in~\Cref{fig:optim_hist_trot}, the algorithm discovers parameters that match the hand-tuned performance in less than 5 minutes (30 trials, a trial takes around $10 \text{s}$ on average, accounting for resets) and parameters that exceed $0.20 \text{m/s}$ in 10 minutes (60 trials).

After only 40 minutes of optimization, the gains are significant for the trotting gait (\cf~\Cref{tab:fast_trot}): with the same maximum motor velocity, the quadruped trots 70\% faster, reaching a mean speed of $0.26 \text{m/s}$ (vs. $0.15 \text{m/s}$ with hand-tuned parameters).

Closing the loop with reinforcement learning helps to further improve performance\footnote{We used bootstrap confidence intervals to test for statistical significance.}. %
Both with the hand-tuned and the optimized CPG, the robot trots significantly faster (20\% and 30\% faster respectively).
The combination of optimized CPG + RL gives the best results ($0.34 \text{m/s}$), and the hand-tuned CPG + RL does not reach the performance of the optimized CPG gait ($0.19 \text{m/s}$).
This is also the fastest walking speed achieved on the \bert platform (the previous record was 0.25 m/s in~\cite{lakatos2018dynamic}).

\begin{table}[ht]
\resizebox{\columnwidth}{!}{%
\begin{tabular}{@{}llllll@{}}
\toprule
                      & \multicolumn{1}{l}{RL}         & \multicolumn{2}{c}{CPG} & \multicolumn{2}{c}{CPG + RL} \\
\multicolumn{1}{l}{} & \multicolumn{1}{l}{} & \multicolumn{1}{l}{hand-tuned} & \multicolumn{1}{l}{optimized} & \multicolumn{1}{l}{hand-tuned} & \multicolumn{1}{l}{optimized} \\ \midrule

\multicolumn{1}{l|}{Speed (m/s) $\uparrow$}  & 0.14 & 0.15  & \textbf{0.26} & 0.19 & \textbf{0.34} \\ \midrule
\multicolumn{1}{l|}{Mean $\ratio$ $\uparrow$} & 1.4 +/- 0.1 & 1.4 +/- 0.1 & \textbf{1.9 +/- 0.1} & 1.6 +/- 0.1 & \textbf{2.0 +/- 0.3} \\ \bottomrule
\end{tabular}
}
\caption{
\footnotesize{
Results for the fast trotting experiment.
The optimized gait trots \textcolor{GithubGreen}{70\%} faster that the hand-tuned controller and exploit more the springs.
RL from scratch merely achieves the hand-tuned CPG performance.
RL on top of CPG further improves the performance.
}
}
\label{tab:fast_trot}
\end{table}

\begin{figure}[ht]
\begin{subfigure}{\linewidth}
	\centering
	\includegraphics[width=\linewidth,trim={0 40 0 0},clip]{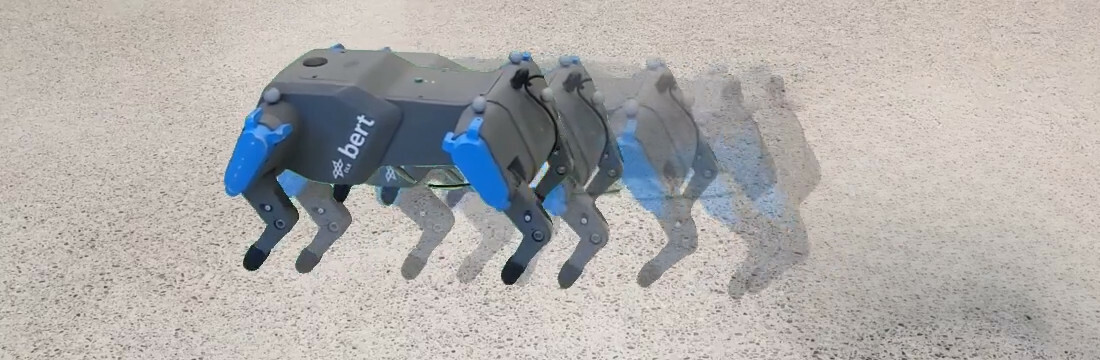}
	\caption{Hand-tuned trotting gait ($0.15 \text{m/s}$).}
	\label{fig:trot}
\end{subfigure}

\begin{subfigure}{\linewidth}
	\centering
	\includegraphics[width=\linewidth,trim={0 20 0 0},clip]{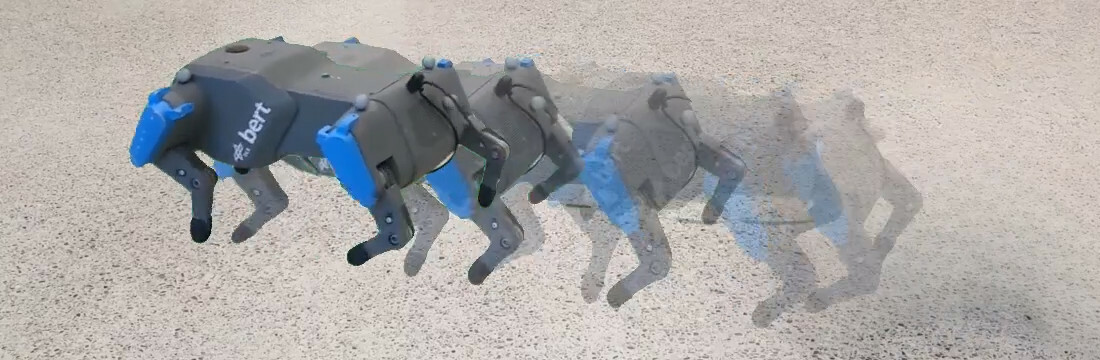}
	\caption{Optimized trotting gait ($0.26 \text{m/s}$).}
	\label{fig:fast_trot}
\end{subfigure}
\caption{Hand-tuned and optimized trotting gaits for a 2-second period.}
\label{fig:trots}
\end{figure}

We display in~\Cref{fig:pattern_trot} the patterns and in~\Cref{fig:trot,fig:fast_trot} the trotting gaits for the hand-tuned and optimized parameters (CPG only).
The main difference is a longer swing phase ($0.29 \text{s}$ vs. $0.14 \text{s}$ for the hand-tuned one) and larger step length ($\textcolor{violet}{\steplen} {=} 5 \text{cm}$ vs. $2.5 \text{cm}$).
However, increasing the step length alone does not result in a faster gait.
Only in combination with appropriate adjustment of the other parameters, the speed can be increased.

\newcommand*\lref[1]{\tikz[baseline=(char.base)]{\node[shape=rectangle,draw,inner sep=2pt] (char) {\scriptsize #1};}}

\begin{figure}[ht]
  \centering
  \multiannotatedimage{
    \includegraphics[width=0.95\linewidth,trim={0 20 0 0},clip]{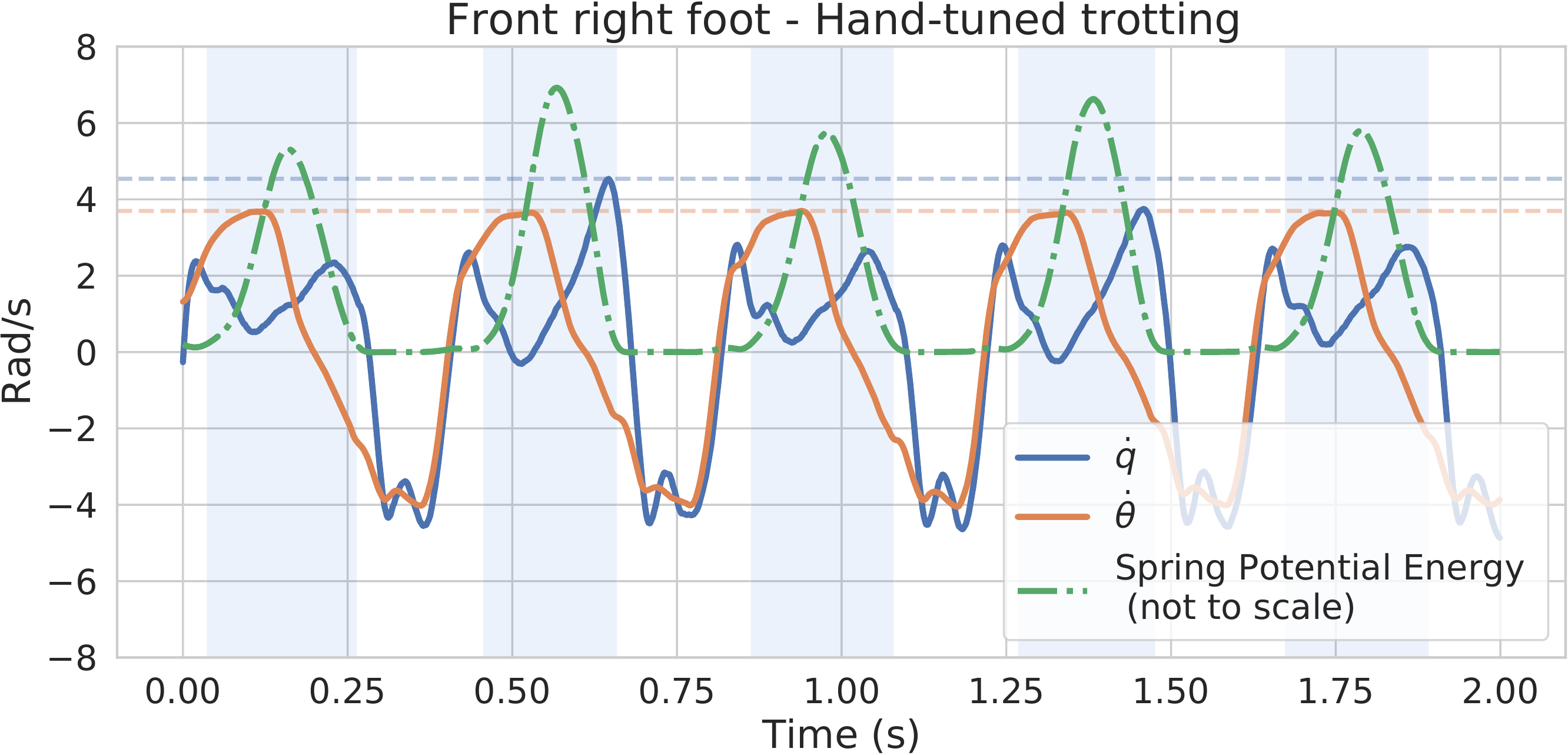}
  }{ %
    \node at (0.387, 0.78) {\tiny A};
  }
  \multiannotatedimage{
    \includegraphics[width=0.95\linewidth]{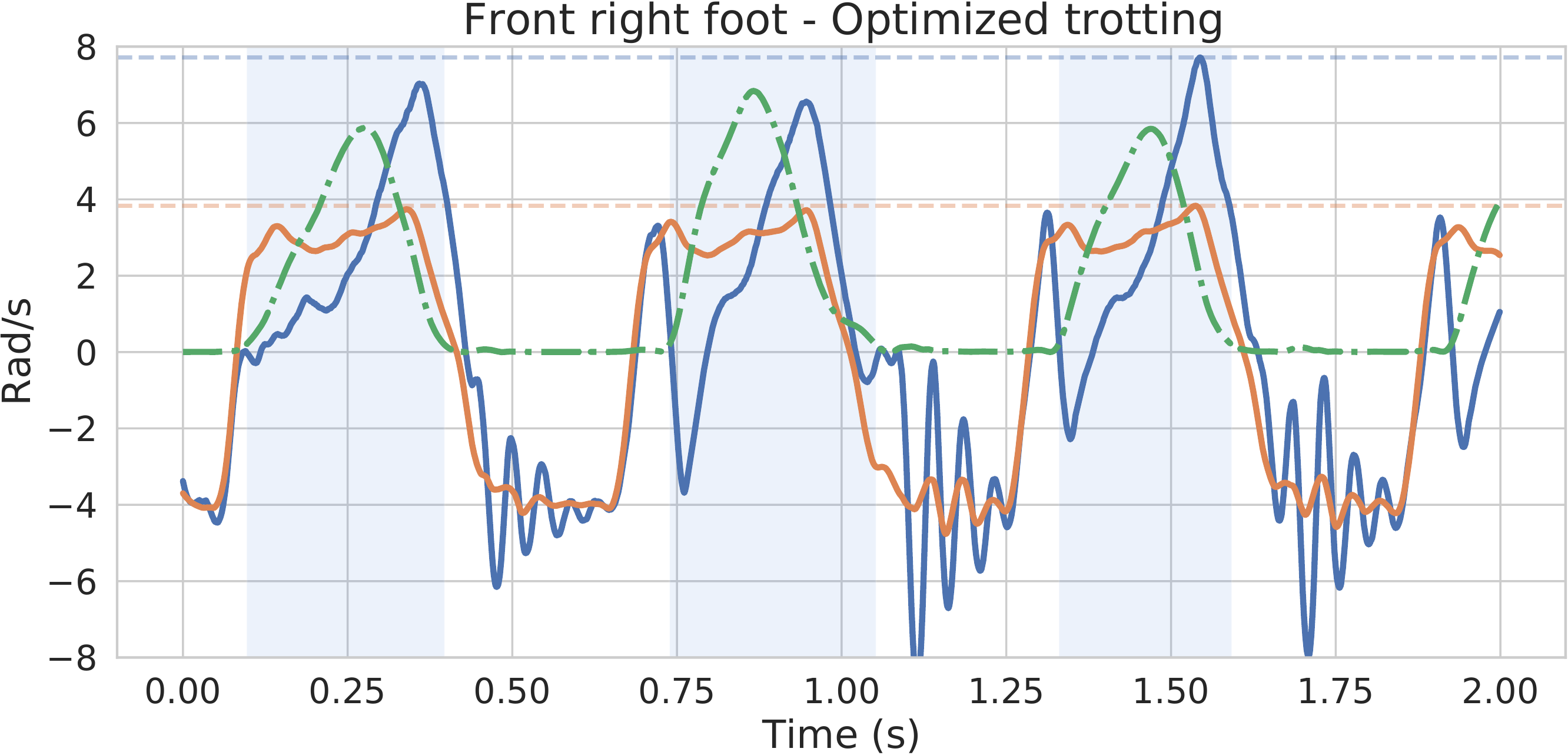}
  }{ %
    \node at (0.51, 0.89) {\tiny B};
  }
\caption{Joint and motor velocity of the front right foot while trotting for a 2-second period.
Darker background represents stance phase.
Blue and orange dotted lines represent maximum joint and motor velocities, respectively.
}
\label{fig:motor_joint_vel}
\end{figure}

Concerning joint and motor velocities, the optimized controller reaches higher peak joint velocities for the same peak motor velocities (\Cref{tab:fast_trot}).
The peak joints velocities are almost twice the one of the motors ($\ratio {=} 2.0$ vs. $1.4$ for the hand-tuned) showing the potential of springs for locomotion.
In fact, as seen in~\Cref{fig:motor_joint_vel}, the peak in joint velocity $\dot{q}$ corresponds to a sudden decrease in the spring potential energy: the energy of the spring is converted into kinetic energy.
The main difference between the results of hand-tuned and optimized parameters is the timing of this conversion.
For the learned controller, it happens when the motor velocity $\motorvel$ is still at its peak~\lref{B}, while for the hand-tuned one, it happens when the motor velocity has already started to decrease~\lref{A}.

\subsection{Is RL from scratch sufficient?}

\Cref{tab:fast_trot} also shows the result of learning to walk from scratch with RL on the real robot.
As shown in the accompanying video, the RL controller learns to walk in 10 minutes only, but the performance plateaus afterward.
The RL controller barely reaches the hand-tuned CPG performance ($0.14 \text{m/s}$), and this translates into less use of the springs ($\ratio \approx 1.4$).

This experiment illustrates the shortcomings of learning from scratch.
RL would require extensive reward engineering~\cite{lee2020learning,miki2022learning} to achieve natural-looking gaits similar to the open-loop CPG (\eg adding a foot clearance reward).
The gait learned by RL is also unpredictable: since we only optimize for the forward speed, it does not have to trot, but could for example, pronk.
On the other hand, the gaits produced by the CPG controller encode basic primitives for walking (\eg foot clearance does not need to be specified in the reward), and the coupling matrix gives control over the gait type.
Learning from scratch with RL is more dangerous for the hardware, as any kind of movement of the legs is allowed, and the smoothness of the controller must be enforced to not damage the motors with high frequency control output.

Finally, we do not consider approaches that train a controller in simulation~\cite{rudin2022learning,miki2022learning}. We want to fully exploit the real robot dynamics and be able to quickly reuse the approach on other robots without the expensive cost of bridging the gap between simulation and reality each time.

\subsection{Stabilizing Pronking}

\begin{figure}[ht]
	\centering
	\includegraphics[width=\linewidth]{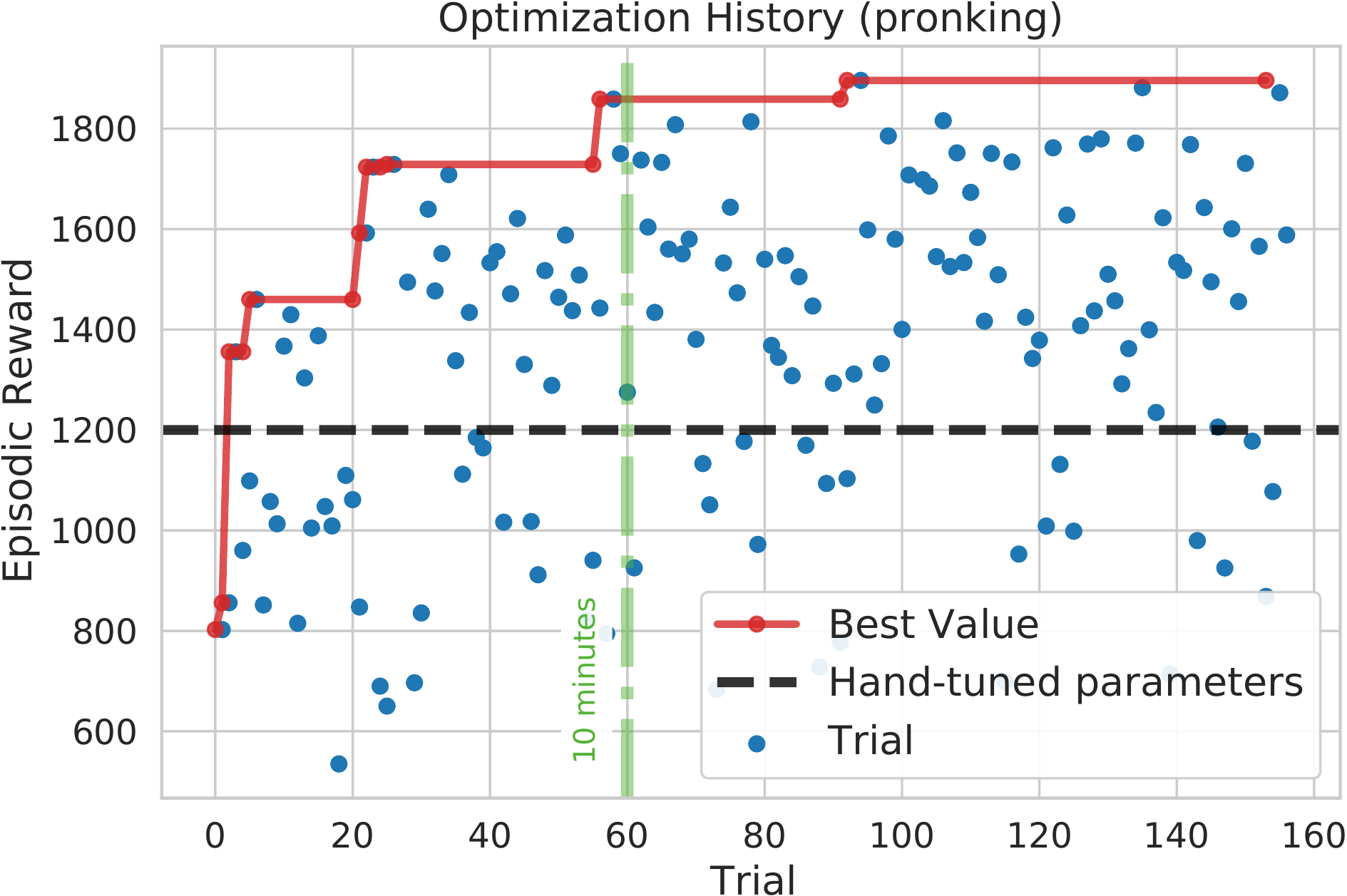}
	\caption{Optimization history plot for pronking (30 minutes).}
	\label{fig:optim_hist_pronk}
\end{figure}

For the pronking task, the BBO algorithm discovers multiple good sets of parameters in less than 10 minutes (60 trials), as seen in the optimization history in~\Cref{fig:optim_hist_pronk}.
The hand-tuned and optimized CPG parameters both allow the robot to jump in place, but the open-loop controller rapidly fails when the robot roll angle increases, either as a consequence of an applied perturbation or the interaction with the floor.

By closing the loop between the robot state and actions, the reinforcement learning controller helps mitigate this issue, as summarized in~\Cref{tab:pronk}.
Here we see that the solution with RL results in no failures.
The qualitative difference is highlighted in the accompanying video.
The quadruped also jumps more in place (less drift with respect to the starting position) while keeping the same jumping height.

As for the trotting experiment, adding RL on top of the hand-tuned CPG improves performance but does not match the optimized controller.

\begin{figure}[ht]
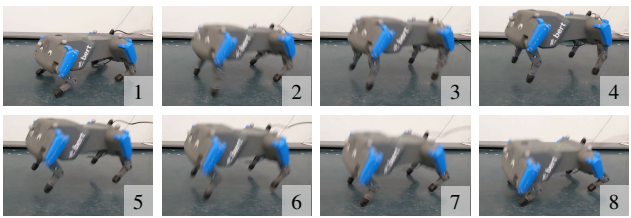

	\centering
  \fourgaitpic{pronk/1.jpg}{1}%
  \fourgaitpic{pronk/3.jpg}{2}%
  \fourgaitpic{pronk/4.jpg}{3}%
	\fourgaitpic{pronk/5.jpg}{4}\\
	\vspace*{0.1cm}
  \fourgaitpic{pronk/7.jpg}{5}%
	\fourgaitpic{pronk/8.jpg}{6}%
	\fourgaitpic{pronk/9.jpg}{7}%
	\fourgaitpic{pronk/10.jpg}{8}%
	\caption{Pronking gait (max jump height: $0.1 \text{m}$).}
	\label{fig:pronk}
\end{figure}

We show in~\Cref{fig:pattern_pronk} the patterns for the hand-tuned and optimized pronking gaits (CPG only), and in~\Cref{fig:pronk} the robot jumping in place.
The optimized parameters are quite different from the hand-tuned ones: the swing phase is shorter ($0.10 \text{s}$ vs. $0.18 \text{s}$ for the hand-tuned one) and stance phase longer ($0.22 \text{s}$ vs. $0.15 \text{s}$), and the ratio between the two is inverted (swing phase shorter than the stance phase).

\begin{figure}[ht]
  \centering
 \multiannotatedimage{
    \includegraphics[width=0.95\linewidth,trim={0 20 0 0},clip]{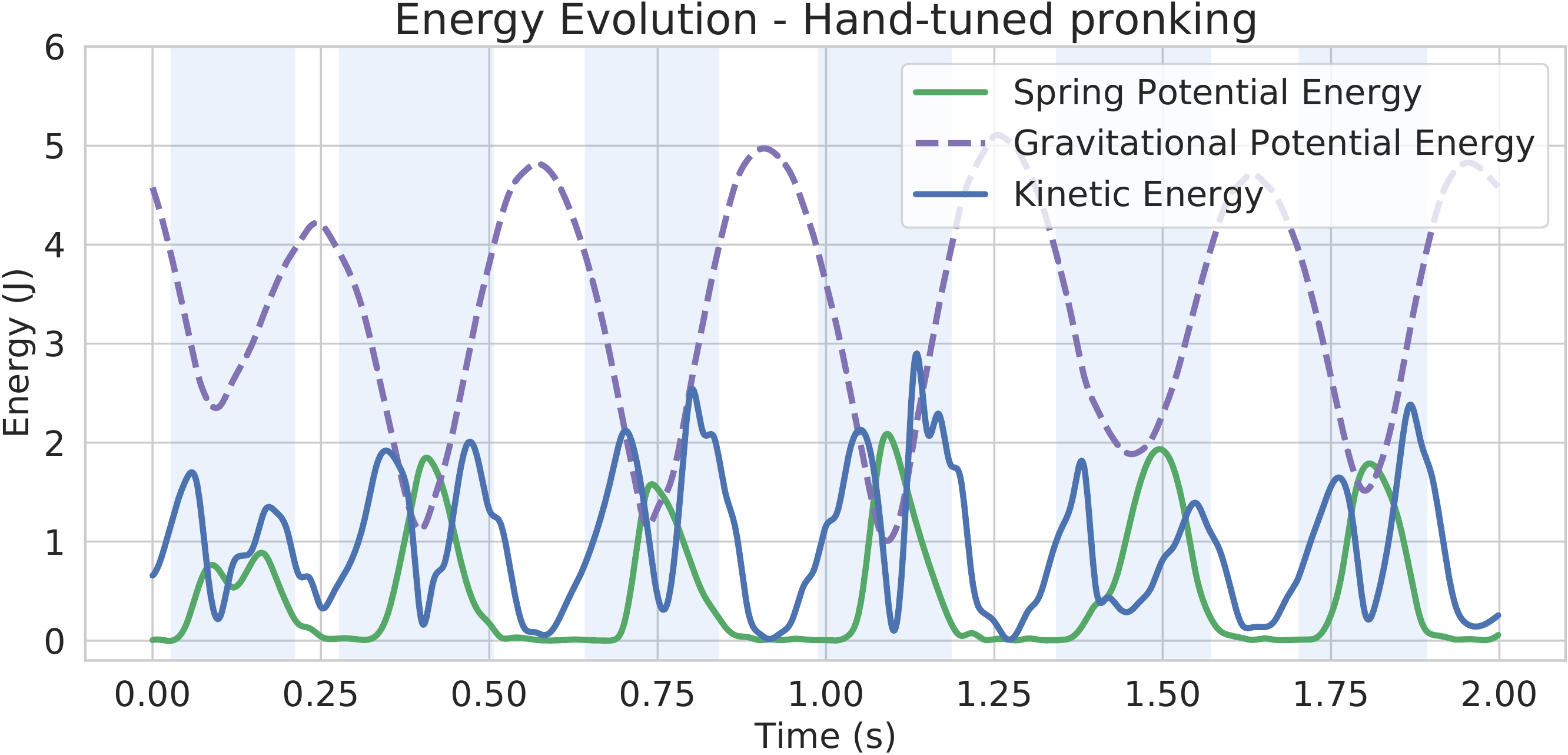}
  }{ %
  }
 \multiannotatedimage{
    \includegraphics[width=0.95\linewidth]{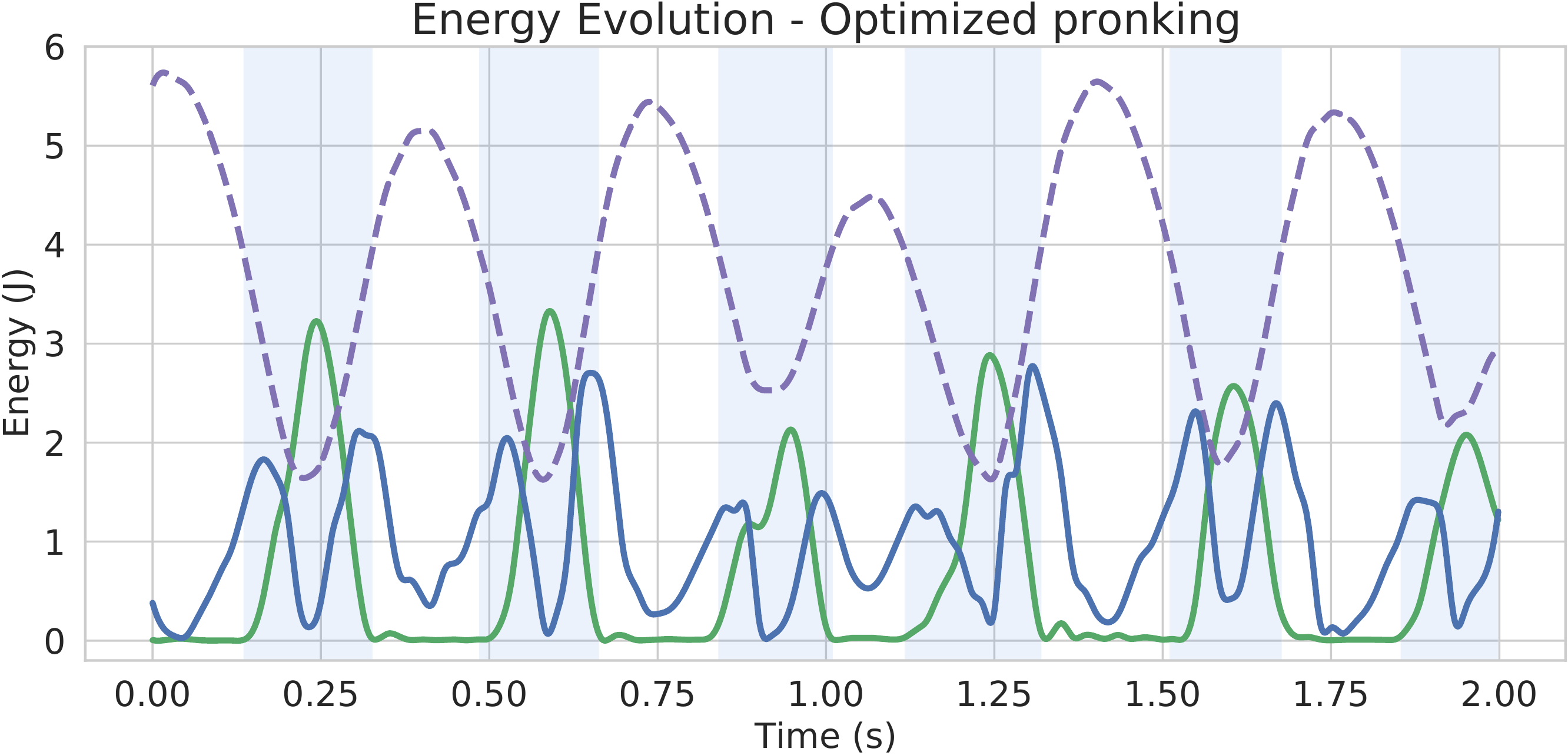}
  }{ %
    \node at (0.348, 0.62) {\tiny C};
		\node at (0.38, 0.54) {\tiny D};
		\node at (0.41, 0.89) {\tiny E};
  }
\caption{Energy evolution while pronking for a 2-second period.}
\label{fig:energy_pronking}
\end{figure}

The energy evolution in~\Cref{fig:energy_pronking} demonstrates how elastic properties of the actuators are exploited.
The robot first crouches on the floor (decrease in gravitational potential energy), compressing and loading the spring~\lref{C}.
To jump, it then pushes all its legs against the floor, while at the same time releasing the energy stored in the springs.
This energy is converted both into kinetic energy (the robot reaches its maximum speed shortly after takeoff~\lref{D}) and gravitational potential energy.
After reaching the maximum height~\lref{E}, the gravitational potential energy is converted back into kinetic energy before the robot lands and the jumping cycle starts again.

The main difference between the hand-tuned and optimized parameters is the maximum spring potential energy.
This results in higher jumps for the optimized controller (higher peak in gravitational potential energy in~\Cref{fig:energy_pronking}).
Finally, compared to trotting, the peak joint velocity is higher for pronking ($\ratio {=} 3.5$ vs. $1.8$) as the leg springs are synchronously compressed and released.
As shown in~\Cref{tab:pronk}, the peak joint velocity is on average $3.5$ the maximum velocity reached by the motors for the optimized parameters, versus $2.9$ for the hand-tuned ones.

\begin{table}[]
\resizebox{\columnwidth}{!}{%

\begin{tabular}{@{}lllll@{}}
\toprule
                                      & \multicolumn{2}{c}{CPG} & \multicolumn{2}{c}{CPG + RL} \\
\multicolumn{1}{l}{} & \multicolumn{1}{l}{hand-tuned} & \multicolumn{1}{l}{optimized} & \multicolumn{1}{l}{hand-tuned} & \multicolumn{1}{l}{optimized} \\ \midrule
\multicolumn{1}{l|}{Mean reward ($10^3$) $\uparrow$}  & 1.0 +/- 0.1 & 1.4 +/- 0.2 & 1.4 +/- 0.1  & \textbf{1.6 +/- 0.1} \textcolor{GithubGreen}{(+60\%)}  \\ \midrule
\multicolumn{1}{l|}{Drift $\drift$ cost $\downarrow$}  & 1.3 +/- 0.1 & 1.5 +/- 0.1 & 1.3 +/- 0.1  & 1.3 +/- 0.1 \\ \midrule
\multicolumn{1}{l|}{Angular vel. cost $\downarrow$}  & 204 +/- 6 & 194 +/- 14 & 183 +/- 10  & \textbf{160 +/- 10} \\ \midrule
\multicolumn{1}{l|}{Max height (cm) $\uparrow$}  & 7.3  & 9.3 & 8.1  & 9.4 \\ \midrule

\multicolumn{1}{l|}{Failures} & \textcolor{GithubRed}{1 failure} & \textcolor{GithubRed}{1 failure}  & \textcolor{GithubGreen}{no failure}  & \textcolor{GithubGreen}{no failure}  \\ \midrule
\multicolumn{1}{l|}{Mean $\ratio$ $\uparrow$} & 2.9 +/- 0.2 & \textbf{3.3 +/- 0.2} & 3.2 +/- 0.2 & \textbf{3.5 +/- 0.3} \textcolor{GithubGreen}{(+20\%)} \\ \bottomrule
\end{tabular}
}

\caption{Results for the jumping in place experiment, on 5 evaluation episodes. Failures occur without external perturbations.
}
\label{tab:pronk}
\end{table}

\section{Conclusion}
We have shown that by integrating central pattern generators with black-box optimization and reinforcement learning, it is possible to learn efficient and natural-looking gaits directly on a real elastic quadruped robot.
This is done without the need for complex reward engineering or massively parallel simulation.
In particular, the learned gaits exploit the natural dynamics of the robot, enabling efficient storage and release of energy in the springs, a behavior that emerged naturally during training without needing to be accounted for in the reward.
This enabled DLR's quadruped \textit{bert} to achieve the fastest walking speed ever recorded.

Learning directly on the real robot comes with its own challenges.
Each experiment must be carefully designed to achieve sample efficient training (choice of action and search space, hyperparameters); trials on the real robot cannot be parallelized.
Like simulation-to-reality approaches, running software on real hardware requires expertise to ensure the robot's safety and is time-consuming.
However, with our approach, once training is complete, no additional step is required for deployment.
Although we focus here on individual gaits, our future research will address the discovery of novel behaviors, especially when adapting to new and unseen environments.

\section*{Acknowledgment}
\label{sec:acknowledgement}
This work was funded by the EU's H2020 Research and Innovation Programme under grant numbers 951992 (VeriDream) and 835284 (M-Runners).
We thank Milan Hermann and Florian Loeffl for their help with the hardware.

\bibliographystyle{./style/IEEEtran}
\bibliography{bibliography}

\end{document}